%% file: root.tex
\documentclass[letterpaper, 10pt, conference]{ieeeconf}

\IEEEoverridecommandlockouts

\usepackage{amsthm}
\usepackage{times}
\usepackage{multicol}
\usepackage[bookmarks=true]{hyperref}
\usepackage{xcolor}
\usepackage{hyperref}
\usepackage{amsmath, amssymb}
\usepackage{amsfonts}
\usepackage{graphicx}
\usepackage{siunitx}
\usepackage{standalone}
\usepackage{booktabs}
\usepackage[ruled,vlined,linesnumbered,noend]{algorithm2e}
\usepackage{mdframed}
\usepackage{fancyvrb,multirow}
\usepackage{soul}
\usepackage{dsfont,mathabx}
\usepackage{array, booktabs}
\usepackage{subfigure}
\usepackage{booktabs}
\usepackage{makecell}
\usepackage{threeparttable}

\theoremstyle{definition}

\newtheorem{remark}{Remark}

\title{\LARGE \bf
HULK: Large-scale Hierarchical Coordination under Continual and Uncertain Temporal Tasks
}

\author{Qingyuan Luo$^1$, Jie Li$^2$ and Meng Guo$^1$
\thanks{
	The authors are with
	$^1$the Department of Mechanics and Engineering Science,
	College of Engineering, Peking University, Beijing 100871, China;
	and $^2$National University of Defense Technology, Hunan 410073, China.
    This work was supported by the National Natural Science Foundation
    of China (NSFC) under grants 62203017, T2121002, U2241214;
	2030-Key Project under Grant 2020AAA0108200;
    and by the Fundamental Research Funds for the central universities.
    Contact: {\tt\small meng.guo@pku.edu.cn}.
}
}

\begin{document}
\maketitle
\thispagestyle{empty}
\pagestyle{empty}

\input{contents/abstract.tex}
\input{contents/introduction.tex}
\input{contents/problem.tex}
\input{contents/solution.tex}

\input{contents/experiment.tex}
\input{contents/conclusion.tex}

\newpage
\bibliographystyle{IEEEtran}
\bibliography{contents/references}

\end{document}

%% file: contents/abstract.tex
\begin{abstract}
Multi-agent systems can be extremely efficient when working concurrently
and collaboratively,
e.g., for delivery, surveillance, search and rescue.
Coordination of such teams often involves two aspects:
(i) selecting appropriate subteams for different tasks in various areas;
(ii) coordinating agents in the subteams to execute the associated subtasks.
Existing work often assumes that the tasks are static and known beforehand,
where an integer program can be formulated and solved offline.
However, in many applications, the team-wise tasks are generated
online continually by external requests;
and the amount of subtasks within each task is uncertain
(e.g., the number of packages to deliver, and victims to rescue).
The aforementioned offline solution becomes inadequate as it would require
constant re-computation for the whole team and global communication
to broadcast the results.
Thus, this work tackles the large-scale coordination problem under continual
and uncertain temporal tasks,
specified as temporal logic formulas over collaborative actions.
The proposed hierarchical framework (HULK)
consists of two interleaved layers:
the rolling assignment of currently-known tasks to sub-teams within a certain horizon,
and the dynamic coordination within a sub-team given the detected subtasks
during online execution.
Thus, the coordination is performed hierarchically at different granularities
and triggering conditions, to improve the computational efficiency and robustness.
It is validated rigorously over large-scale heterogeneous systems under
various temporal tasks and environment uncertainties.
\end{abstract}

%% file: contents/introduction.tex
\section{Introduction}\label{sec:intro}
Fleets of heterogeneous robots,
such as ground vehicles and aerial vehicles,
are deployed to accomplish tasks that are otherwise too inefficient or even infeasible
for a single robot~\cite{arai2002advances}.
Not only the overall efficiency of the team can be significantly improved
by allowing the robots to move and act concurrently~\cite{toth2002overview, cliff2015online};
but also the capabilities of the team can be greatly extended by
enabling multiple robots to directly collaborate on a task~\cite{fink2008multi, varava2017herding}.
However, the optimal coordination of a large-scale multi-agent system
to accomplish the desired task is well-known to be hard,
especially to fulfill the spatially distributed subtasks in the right order at the right time.
The set of possible task assignments are often combinatorial with respect to the number of robots
and the length of tasks~\cite{arai2002advances, toth2002overview, kantaros2020stylus}.
Commonly such a team-wise task is specified beforehand and remains {unchanged},
of which the solutions are derived {offline} and thus {static}.
A particularly challenging scenario is when the system operates indefinitely,
i.e., new tasks are released or canceled \emph{dynamically} and \emph{continually}
by external demand,
thus requiring the agents to change their task plans frequently.
The aforementioned methods become inadequate as the sequence of tasks is infinite
and their specifications are unknown beforehand.
Recursive application of the static methods in a naive way leads to
not only intractable computation complexity,
but also inconsistent or even oscillatory assignments.

\subsection{Related Work}\label{subsec:intro-related}
\begin{figure}[t]
  \centering
  \includegraphics[width=0.95\hsize]{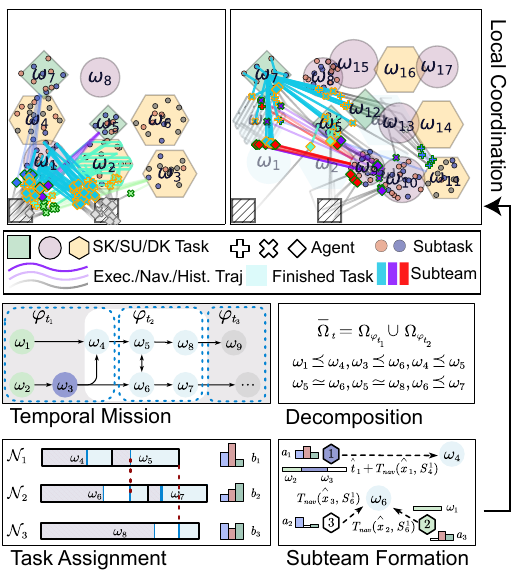}
  \vspace{-0.1in}
  \caption{Overall framework, including:
    snapshots of online execution of \textbf{three} types of tasks,
    which contains numerous subtasks (\textbf{top});
    the posets associated with missions that are released online (\textbf{middle});
    and the receding-horizon assignment of tasks to optimized formation of teams (\textbf{bottom}).
  }
  \label{fig:overall}
  \vspace{-0.2in}
\end{figure}

Task planning refers to the process of first decomposing this task into sub-tasks
and then assigning them to the team,
see~\cite{torreno2017cooperative,gini2017multi, khamis2015multi} for comprehensive surveys.
Different optimization criteria can be chosen, such as
MinSUM~\cite{gini2017multi};
and MinMAX~\cite{nunes2015multi}.
The tasks can be specified in various forms, such as
in the multi-vehicle routing
problem~\cite{khamis2015multi},
the job-shop problem~\cite{brucker1994branch};
and the coalition formation
problem~\cite{massin2017coalition}.
Existing methods can be categorized into centralized methods such as
mixed integer linear programming (MILP)~\cite{torreno2017cooperative}
and search-based methods~\cite{fukasawa2006robust};
and decentralized methods such as
market-based methods~\cite{luo2015distributed} and distributed
constraint optimization (DCOP)~\cite{boerkoel2013distributed}.
However, since many task planning problems are in general NP-hard or even
NP-complete~\cite{gini2017multi},
meta-heuristic approaches are used to gain computational efficiency,
e.g., local search~\cite{hoos2004stochastic} and genetic algorithms~\cite{khamis2015multi}.
However, most of the aforementioned methods can not be applied directly in
this work due to the general task specification as temporal logic
formulas over collaborative actions.
Additionally, the aforementioned methods often focus on solving a static
problem, rather than the continual and dynamic scenario addressed in this work.

Temporal logic formulas can be used to specify complex robotic tasks,
such as Probabilistic Computation Tree Logic (PCTL) in~\cite{lahijanian2011temporal},
Linear Temporal Logics (LTL) in~\cite{kantaros2020stylus, schillinger2018simultaneous,
luo2021temporal,jones2019scratchs, guo2015multi, tumova2016multi},
and counting LTL (cLTL) in~\cite{sahin2019multirobot}.
Considerable results are developed in the recent years regarding the task assignment problem
of team-wise tasks specified as temporal logic formulas.
Analogously, they can be categorized into centralized methods and decentralized methods.
Centralized methods often put emphases on optimality and completeness,
such as the sampling-based search algorithm~\cite{kantaros2020stylus},
the simultaneous decomposition and assignment method~\cite{schillinger2018simultaneous},
the MILP formulation~\cite{luo2021temporal, sahin2019multirobot,
jones2019scratchs}.
Decentralized methods are more applicable to large-scale multi-agent systems,
including the local coordination of local
tasks~\cite{guo2015multi, tumova2016multi, guo2016task},
the local assignment under partial workspace~\cite{menghi2018multi},
the distributed sampling method~\cite{kantaros2018distributed},
and the online auction algorithm~\cite{schillinger2019specification}.
However, the scenario where new tasks are released online with uncertain subtasks
is rarely addressed in the aforementioned work,
as it would require online adaptation for both the global task assignment
and the local subtask execution.

\subsection{Our Method}\label{subsec:intro-our}
To tackle these issues, this work put strong emphases on the online coordination algorithm
that are essential when the collaborative tasks are released continually,
and the amount of subtasks within each task is uncertain.
In other words, both the distribution and requirements of the tasks
can only be known during online execution.
As shown in Fig.~\ref{fig:overall},
the proposed method is based on the hierarchical coordination framework (HULK)
that combines the global task assignment and the local subtask coordination.
Namely, the global mission specification is first decomposed into collaborative
tasks with the associated temporal constraints.
Then, a receding-horizon assignment algorithm is applied
to assign these collaborative tasks to subteams of the agents,
subject to resources requirements, navigation cost and ordering constraints.
Afterwards, each subteam follows different local strategies,
where subtasks are detected and assigned dynamically during execution.
Online adaptation at both levels is triggered by external events
and execution status.
Efficiency and robustness of the proposed framework
is validated rigorously over large-scale heterogeneous systems
and three different temporal tasks.

Main contribution of this work is threefold:
(I) The proposed hierarchical coordination algorithm is applicable
to a wide range of formulations, such as heterogeneous agents,
temporal tasks and collaborative actions;
(II) It is robust to varying distribution of continual tasks,
contingent agent failures and uncertain subtasks;
and (III) It is computationally efficient and scalable to large-scale systems.

%% file: contents/problem.tex
\section{Problem Description}\label{sec:problem}
\subsection{Multi-agent Systems}\label{subsec:multi-agent}
Consider a team of~$N$ agents that share the common
workspace~$\mathcal{W}\subset \mathbb{R}^3$.
Each agent~$i\in \mathcal{N}\triangleq \{1,\cdots,N\}$ is described by its position~$x_i\in \mathcal{W}_i$
and its action~$a_i\in \mathcal{A}_i$,
where~$\mathcal{W}_i\subseteq \mathcal{W}$ is the allowed workspace;
and~$\mathcal{A}_i$ is the set of primitive actions
such as surveillance, delivery, capture and defense.
Each agent can navigate freely in the workspace via
a reference velocity~$v_i\in \mathbb{R}^3$.
Thus, the local plan of an agent is given by a sequence of
timed goal positions and performed actions,
i.e.,~$\tau_i\triangleq (t^1_i,\,g^1_i,\,a^1_i)
(t^2_i,\,g^2_i,\,a^2_i)\cdots$
with~$t^\ell_i\geq 0$,~$g^\ell_i \in \mathcal{W}_i$,~$a_i^\ell \in \mathcal{A}_i$ being
the time instant, goal position and action,
$\forall \ell\geq 1$.
In other words, under this local plan,
agent~$i\in \mathcal{N}$ should navigate to~$g_i^\ell$ with velocity $v_i$
and start performing~$a_i^\ell$ from time~$t_i^\ell$, for all~$\ell\geq 1$.

\subsection{Mission Specifications}\label{subsec:mission_spec}
At time~$t\geq 0$, a temporal mission is released (e.g., by a human operator
or triggered by an event) to the fleet, denoted by~$\varphi_t\triangleq \text{sc-LTL}
\big(\boldsymbol{\omega}_t\big)$,
where: (I) $\boldsymbol{\omega}_t\triangleq \{\omega_1,\cdots,\omega_{M_t}\}$
is the set of collaborative tasks within the mission.
Each collaborative task~$\omega_m\in \boldsymbol{\omega}_t$ is denoted by:
\begin{equation}\label{eq:collab_task}
\omega_m\triangleq \Big(S_m,\, \eta_m, \,\big\{\big(n_j,\,a_j,\,s_j\big),
j=1,\cdots, J_m\big\}\Big),
\end{equation}
where~$S_m\subset \mathcal{W}$ is the area
that the task should be accomplished;
$(n_j,a_j,s_j)$ is a subtask that requires
\emph{at least}~$n_j$ agents performing action~$a_j$ collaboratively
at location $s_j\in S_m$;
$J_m>0$ is the total number of subtasks within the task~$\omega_m$;
and the estimated duration of each subtask is given by
function~$\eta_m:\mathbb{N} \times \mathcal{A}
\times 2^{\mathcal{N}} \rightarrow \mathbb{R}^+$, i.e.,
$\eta_m(n_j,\,a_j,\, \mathcal{N}_j)$ returns
the duration if the subteam of agents~$\mathcal{N}_j\subset \mathcal{N}$
is assigned to provide the required action~$a_j$ at location~$s_j$.
Thus, task~$\omega_m$ is accomplished after each subtask is completed;
(II) The tasks~$\omega_m$ are nested following the syntax of
Linear Temporal Logic (LTL)~\cite{baier2008principles},
e.g., via $\varphi \triangleq \top \;|\; p  \;|\; \varphi_1 \wedge
\varphi_2  \;|\; \neg \varphi  \;|\; \bigcirc \varphi  \;|\;  \varphi_1 \,\textsf{U}\, \varphi_2,$
where $\top\triangleq \texttt{True}$, $p \in AP$, $\bigcirc$ (\emph{next}),
$\textsf{U}$ (\emph{until}) and $\bot\triangleq \neg \top$.
or other derived operators like $\Box$ (\emph{always}),
$\Diamond$ (\emph{eventually}), $\Rightarrow$ (\emph{implication}).
The full semantics and syntax of syntactic co-safe LTL (sc-LTL) are omitted here for brevity,
see e.g.,~\cite{baier2008principles}.

Consequently, the temporal mission~$\varphi_t$ is satisfied if
all collaborative tasks are accomplished,
while the resulting trace of their temporal ordering
satisfies the sc-LTL formula~$\varphi_t$ via
the satisfaction relation~$\models$ from~\cite{baier2008principles}.
Lastly, the accumulated missions up to time~$t\geq 0$ are denoted
by~$\boldsymbol{\varphi}_t\triangleq
\{{\varphi}_{t_\ell}, \, \forall t_\ell \leq t\}$.
It is worth mentioning that the tasks~$\{\omega_m\}$ in~\eqref{eq:collab_task}
can be \emph{uncertain}, i.e., the exact number of subtasks~$J_m$
and their locations $\{s_j\}$ are unknown at the time of release.
This might be due to the partial observability or dynamic nature
of the environment.

\begin{remark}\label{rm:cltl}
  The definition of collaborative task in~\eqref{eq:collab_task}
  differs from the notion of cLTL~\cite{sahin2019multirobot} in two aspects:
  (I) Both the location and number of subtasks are uncertain;
  (II) The duration of all subtasks can vary depending on the
  assigned agents, i.e., instead of being instantaneous~\cite{liu2024time}.
  \hfill $\blacksquare$
\end{remark}

\begin{remark}\label{rm:constraints}
  The duration function~$\eta_m(\cdot)$ typically \emph{saturates}
  as the number of participants increases,
  i.e., the marginal benefits diminishes
  as also adopted in~\cite{apt2009generic,chen2024accelerated}.
  \hfill $\blacksquare$
\end{remark}

\subsection{Problem Statement}\label{subsec:prob}
Given the above model, the overall objective is to synthesize the
collective plans~$\{\tau_i\}$ such that the average response
of each mission is minimized, i.e.,
\begin{equation}\label{eq:objective}
  \bold{min}_{\{\tau_i\}}\,
  \frac{\sum_{{\varphi}_{t_\ell} \in {\boldsymbol{\varphi}}_t}
  (t_\ell^{\texttt{f}}-t_{\ell})}
  {|{\boldsymbol{\varphi}}_t|},
\end{equation}
where~$0\leq t_\ell\leq t^{\texttt{f}}_{\ell}$ are the time instants
when the mission~${\varphi}_{t_\ell} \in {\boldsymbol{\varphi}}_t$
is released and accomplished, respectively.

%% file: contents/solution.tex
\section{Proposed Solution}\label{sec:solution}
The proposed solution consists of two main components:
(I) The receding-horizon task planning algorithm that assigns tasks
to subteams of agents given the global mission specification
and constraints on the resources;
(II) The local coordination algorithm that assigns subtasks to agents
during online execution.
The synergy and adaptation of both components are triggered
by external events and execution status.
\input{contents/task.tex}
\input{contents/act.tex}
\input{contents/overall.tex}

%% file: contents/task.tex
\subsection{Receding-horizon Task Assignment}\label{subsec:task_assignment}
\subsubsection{Decomposition of Temporal Tasks}\label{subsec:partial}
Given the mission specification~$\varphi$,
the NBA associated with~$\varphi$ is denoted by~$\mathcal{B}=(Q,\,Q_0,\,\Sigma,\,\delta,\,Q_F)$,
of which the notation follows~\cite{baier2008principles}.
Based on~$\mathcal{B}$, the associated tasks and their partial
temporal constraints can be computed based on our earlier work~\cite{liu2024time, liu2024fast},
as posets over tasks, i.e.,
\begin{equation}\label{eq:task_posets}
\Omega_\varphi \triangleq \big{\{}(\Omega,\,\preceq,\,\simeq)\big{\}},
\end{equation}
where~$\Omega\triangleq \{\omega_1, \cdots, \omega_M\} \subset \Sigma$ is a set of tasks;
the partial ordering constraints~$\preceq,\,\simeq \subset \Omega \times \Omega$
such that: (I) $\omega_{m_1} \preceq \omega_{m_2}$ if task~$\omega_{m_1}$ should be
accomplished {before} task~$\omega_{m_2}$ is started;
(II) $\omega_{m_1} \simeq \omega_{m_2}$ if tasks~$\omega_{m_1}$ and~$\omega_{m_2}$
should start {at the same time}.
Simply speaking, the posets are abstracted from the mission automaton as the set
of possible ways to satisfy the mission, as the set of tasks involved and their
temporal relations. This allows more parallel execution during assignment thus
improving efficiency.
Detailed algorithms and completeness analyses can be found in~\cite{liu2024time}.

As the missions can be specified online,
the set of missions by time~$t>0$ is given by~$\boldsymbol{\varphi}_t$,
each of which can be decomposed into the partially-ordered tasks.
Consequently, the set of all \emph{known and unfinished} tasks by time~$t$ is given by:
$\overline{\Omega}_t\triangleq \bigcup_{\varphi_i\in \boldsymbol{\varphi}_t}\,\Omega_{\varphi_i}$,
where~$\Omega_{\varphi_i}$ is the partially-ordered tasks
associated with mission~$\varphi_i \in \boldsymbol{\varphi}_t$ as defined in~\eqref{eq:task_posets}.
Note that the tasks from different missions are assumed to be independent.
Moreover, given~$\overline{\Omega}_t$, a directed acyclic \emph{task graph}~$\mathcal{G}_t$
can be constructed for these tasks based on their partial ordering,
i.e., each node in the graph represents a task in $\overline{\Omega}_t$,
and the edges represent the precedence and concurrence constraints.

\subsubsection{Capacity-based Task Assignment for Subteams}\label{subsec:assignment}

Given the temporal constraints on simultaneous execution and
the objective to minimize response time,
the agents need to be divided into subteams for parallel execution.
However, due to limitations in the number and capacity of the agents,
it is not feasible to create a subteam for each individual task,
rather a subteam is assigned a sequence of tasks
based on agent capacity and the requirements of the tasks.

Denote by~$\nu \triangleq \{\mathcal{C}_k,\, k = 1, \cdots, K\}$
the set of partial assignments along with the capacity constraints for each subteam,
where each local assignment is given by:
\begin{equation} \label{eq:init-subteam}
  \mathcal{C}_k \triangleq \Big( \big(\omega_k^0, \cdots, \omega_k^{L_k}\big),
  \big\{(m^j_k,\, a^j_k),\, a^j_k \in \mathcal{A}\big\} \Big),
\end{equation}
of which the first part is the sequence of tasks assigned to the $k$-th subteam,
$L_k$ is the total number of tasks in the sequence,
while the second is the minimum number of agents~$m^j_k$
to perform action~$a^j_k$.
Note that the number of subteams~$K$ is not pre-defined and to be optimized.

\begin{algorithm}[t!]
  \caption{Task Assign. and Subteam Format.}
  \label{alg:task}
  \SetAlgoLined
  \KwIn{Robots~$\mathcal{N}$, tasks~$\boldsymbol{\varphi}_t$, horizon~$H$.}
  \KwOut{Assignment~$\{\nu_{K^\star}^\star\}$, subteams~$\{\mathcal{N}_k\}$,
    and local plans~$\{\xi_i\}$.}
  \tcc{\textbf{Task Assignment}}
  Compute posets~$\overline{\Omega}_t$ for~$\boldsymbol{\varphi}_t$ by~\eqref{eq:task_posets}\;
  Build task graph~$\mathcal{G}_t$ given~$\overline{\Omega}_t$\;
  Initialize~$\mathcal{V}^\star=\emptyset$\;
  \For{$K \in \mathcal{H}$}
  {
    Compute optimal assignment~$\nu^\star_{K}$ via~\eqref{eq:node-makespan}\;
    Add~$\nu^\star_{K}$ to~$\mathcal{V}^\star$\;
  }
  Choose best~$\nu^\star_{K^\star}$ among~$\mathcal{V}^\star$ by~\eqref{eq:node-makespan}\;
  \tcc{\textbf{Subteam Formation}}
  Compute cost matrix~$\{t_{ik}\}$ by~\eqref{eq:robot-task-time}\;
  Formulate and solve MILP problem to find~$\{b_{ik}\}$\;
  Compute local plans~$\{\xi_i\}$ by~\eqref{eq:team-plan}\;
\end{algorithm}

Consider that~$H>0$ tasks should be assigned within the task graph~$\mathcal{G}_t$.
A search-based algorithm is proposed to determine the optimal subteam assignment~$\nu^\star$.
As summarized in Alg.~\ref{alg:task},
starting from the root node as the empty assignment~$\nu_0$,
the selected node is expanded by adding the next feasible task to any of the subteam.
In particular, the assignment for the set of tasks
that are currently being executed remains unchanged,
while the rest of the tasks within~$\mathcal{G}_t$ can be
added if their preceding tasks are fulfilled or being executed.
More importantly, after adding~$\omega^{L_k}_k$ to the subteam~$k$,
the capacity constraint is updated as follows:
\begin{equation}\label{eq:update-constraint}
  m^j_k \triangleq \underset{a^j_k\in \omega^\ell_k\in \mathcal{C}_k}{\textbf{max}}
  \{n_j\},\;  \forall a^j_k\in \mathcal{A};
\end{equation}
i.e., the maximum number of agents~$m^j_k$ required
for each action~$a^j_k$ across the sequence of task~$ \omega^\ell_k$.
The search is terminated when the number of assigned tasks for the fleet reaches the horizon~$H$,
or the capacity constraints are violated, i.e.,
\begin{equation}\label{eq:capacity-bound}
  \sum_{k\in \mathcal{K}} m^j_k \leq
  \sum_{i\in \mathcal{N}}{\mathds{1}(a^j_k \in \mathcal{A}_i)},\; \forall a^j_k \in \mathcal{A};
\end{equation}
where the left-side is the resources required by the assignment,
and the right-side is the complete capacity of the fleet.
Lastly, the optimal assignment is selected from the complete tree
by evaluating the overall quality of each node, i.e.,
\begin{equation}\label{eq:node-makespan}
  \begin{split}
    \xi(\nu) &\triangleq \eta_{L_k}(\nu,\, \overline{\Omega}_t) +
    \underset{k\in \mathcal{K}}{\textbf{max}}\,
    \big{\{}t_{\texttt{e}}(\omega_k^{L_k})\big{\}};\\
    t_{\texttt{e}}(\omega_k^\ell) & \triangleq
    \Big{(}\underset{\omega_j \in
    \texttt{Pre}({\omega_k^\ell})}{\textbf{max}}
    \big{\{}t_{\texttt{e}}(\omega_j)\big{\}} \Big{)}
    + T_{\texttt{nav}}(S_k^{\ell-1}, S_k^{\ell}),
\end{split}
\end{equation}
where~$t_{\texttt{e}}(\omega^\ell_k)$ is the estimated
ending time of~$\omega^\ell_k \in \mathcal{C}_k$;
$\texttt{Pre}({\omega_k^\ell})$ is set of preceding tasks in the task graph~$\mathcal{G}_t$;
$T_{\texttt{nav}}(S_k^{\ell-1}, S_k^{\ell})$ is the estimated navigation time
from the previous task region~$S_k^{\ell-1}$ to the current task region~$S_k^{\ell}$;
and~$\eta(\nu,\, \Omega_t)$ is the estimated progress achieved by the assignment~$\nu$
w.r.t. the unfinished tasks.
Denote by~$\nu^\star_K$ the optimal assignment for~$K$ subteams.
The same search procedure is repeated for different choices of~$K\in \mathcal{H} \triangleq \{1,\cdots, H\}$,
for which the set of optimal assignments is given by~$\mathcal{V}^\star \triangleq \{\nu^\star_K,\, K\in \mathcal{H}\}$.
Within this set, the same measure as in~\eqref{eq:node-makespan} is adopted to select the best choice of~$K$ as~$K^\star$ and the associated assignment~$\nu^\star_{K^\star}$.
It is worth noting that the subteams~$\{\mathcal{C}_k\}\in \nu^\star_{K^\star}$
only specify the constraints on capacity, rather than specific agents.

\begin{remark}\label{remark:prune}
Note that various heuristics can be applied to prune the search space, e.g.,
elimination of symmetric nodes; removing branches if its root node is worse
than the current best node.
More numeric details are given in the Sec.~\ref{sec:experiments}.
\hfill $\blacksquare$
\end{remark}

\subsubsection{Redundancy-aware Subteam Formation}\label{subsec:group}
Given the optimal task assignment for subteams~$\nu_{K^\star}^\star$,
the actual formation of each subteam should be determined, i.e.,
to find the set of subteams~$\overline{\mathcal{N}} \triangleq \{\mathcal{N}_1,\cdots,\mathcal{N}_{K^\star}\}$,
where~$\mathcal{N}_{k_1} \bigcap \mathcal{N}_{k_2}=\emptyset$,
$\forall k_1 \neq k_2$ and $\mathcal{N}_{k_1}, \mathcal{N}_{k_2}\subset \mathcal{N}$.
In other words, the robots in~$\mathcal{N}_k$ are assigned to the subteam~$\mathcal{C}_k$
within~$\nu_{K^\star}^\star$.

This can be done in two steps:
(I) The lower and upper bound for each task is determined by the capacity constraints;
(II) A constrained min-max assignment problem is formulated for the~$N$ robots and the first task of~$K$ subteams,
i.e., the estimated starting time if robot~$i\in \mathcal{N}$ participates in the first
task of
subteam~$\mathcal{C}_k$ is given by:
\begin{equation}\label{eq:robot-task-time}
 t_{ik} \triangleq \widehat{t}_i + T_{\texttt{nav}}(\widehat{x}_i,\, S_k^{1}),
\end{equation}
where~$\widehat{t}_i$ and $\widehat{x}_i$ are the expected time and position when
robot~$i$ becomes available after executing its current task (if any);
$T_{\texttt{nav}}$ is the estimated duration to navigate to the task region~$S_k^{1}$.
It can be formulated as a mixed-integer linear programming (MILP) problem
over the possible robot-task pairings with boolean variables~$\{b_{ik}\}$,
which can be solved efficiently by existing solvers such as \texttt{GLOP}~\cite{glop}.
Thus, the local \emph{task plan} of each robot~$i\in \mathcal{N}_k$ is given by:
\begin{equation}\label{eq:team-plan}
\xi_i\triangleq (S^1_k,\, \omega^1_k) (S^2_k,\, \omega^2_k)
\cdots (S^{L_k}_k,\, \omega^{L_k}_k),\; \forall i \in \mathcal{N}_k;
\end{equation}
as a timed sequence of tasks,
where~$\omega^\ell_k\in \Omega$ is the assigned~$\ell$-th task of
subteam~$\mathcal{N}_k\in \overline{\mathcal{N}}$;
$S^\ell_k\subset \mathcal{W}$ is the associated region;
and~$L_k>0$ is the total length.
Note that the total number of tasks~$\overline{\Omega}_t$
is much larger than the number of subteams,
and continually expanded online.

%% file: contents/act.tex
\subsection{Local Task Coordination}\label{subsec:act}
\begin{figure}[t]
    \centering
    \includegraphics[width=0.95\hsize]{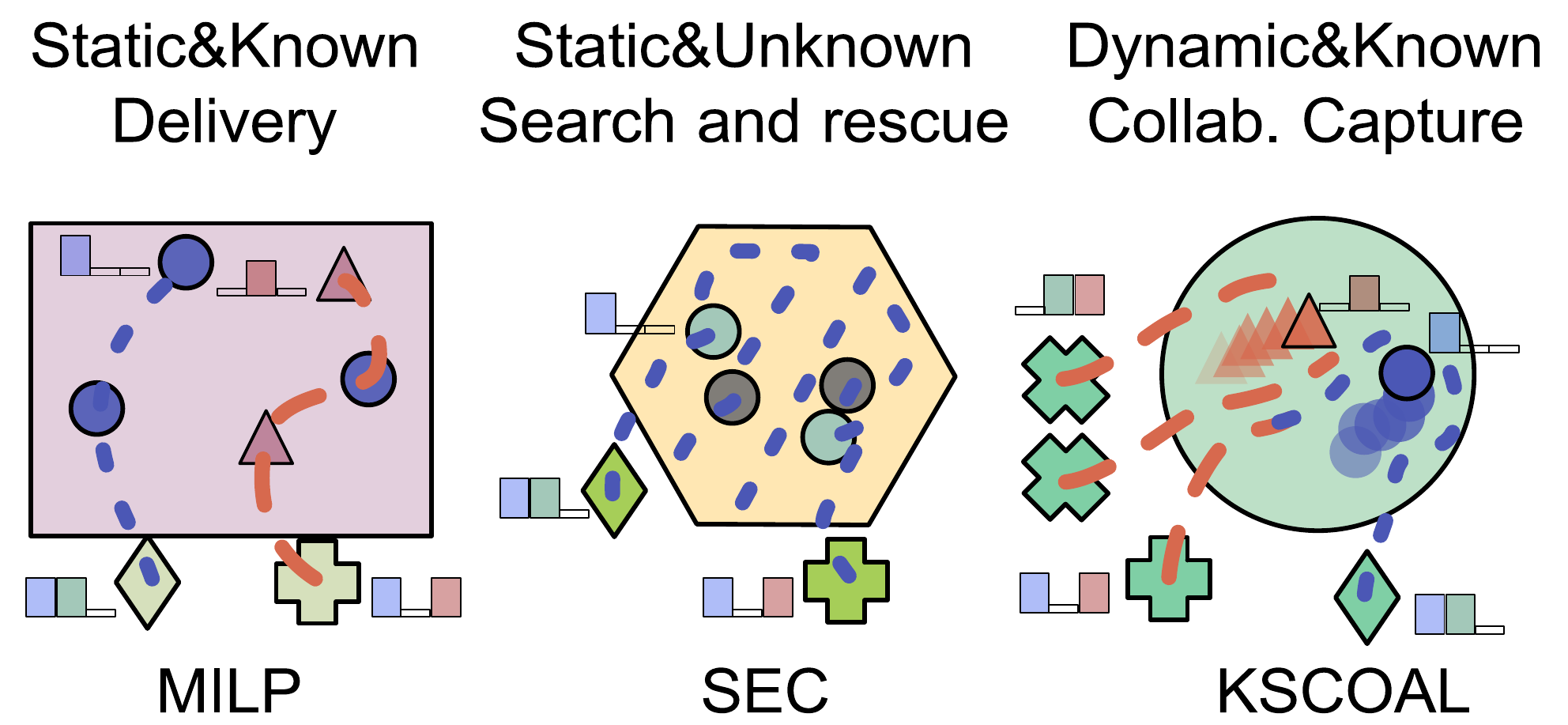}
    \vspace{-0.1in}
    \caption{Illustration of three types of local tasks described in
      Sec.~\ref{subsec:act},
    and the associated coordination strategy.}\label{fig:coordination}
    \vspace{-0.1in}
  \end{figure}

Once the local task plans are derived,
each robot~$i\in \mathcal{N}$
starts executing its $\ell$-th task~$(S^\ell_k,\, \omega^\ell_k)$, i.e.,
to navigate to region~$S^\ell_k$ and perform the task~$\omega^\ell_k$.
The required subtasks from~\eqref{eq:collab_task} are denoted by~$\mathcal{J}^\ell_k\triangleq
\big\{(n_j,\,a_j,\,s_j),\, j=1,\cdots,J^\ell_k\big\}$.
Consequently, all agents in~$\mathcal{N}_k$ should collaboratively fulfill
these subtasks to minimize the overall duration.
More specifically, the local \emph{action plan} of each agent~$i\in \mathcal{N}_k$
for task~${\omega}^\ell_k$ is given
by~$\tau_i=(t^1_i,\,g^1_i,\,a^1_i)(t^2_i,\,g^2_i,\,a^2_i)\cdots$,
as the sequence of timed goal, positions and actions.
The overall objective of the local coordination is to optimize
the collective plans~$\boldsymbol{\tau}^\ell_k \triangleq \{\tau_i,
\,i\in \mathcal{N}_k\}$
such that the makespan of~${\omega}^\ell_k$, denoted by~$T^\ell_k$,
is minimized, i.e., $\textbf{min}_{\boldsymbol{\tau}^\ell_k}\, \{T^\ell_k\}$.
However, as specified earlier, there are uncertainties
regarding the total number of subtasks~$J^\ell_k$ and their locations~$\{s_j\}$
within the task~$\omega^\ell_k$.
Thus, depending on the characteristics of the tasks and workspace,
\emph{\textbf{three}} different local coordination strategies are adopted,
as shown in Fig.~\ref{fig:coordination}.

\subsubsection{Static and Known Tasks}
As the first case, consider that the locations and the number of subtasks
are all known and static.
For instance, the  task of ``delivery'' often consists of several
locations to visit in a region and collaboratively deliver some objects,
which are often known beforehand according to orders.
In this case, variants of the multi-vehicle routing problem can be formulated
as a MILP by enforcing the collaborative actions
at each location.
The key is that the constraints are formulated according to the navigation model of each
agent
and the duration function~$\eta^\ell_k$ from~\eqref{eq:collab_task} given the assignment variables.
The exact formulation is omitted here due to limited space.
Denote by~$\boldsymbol{\tau}^{\ell,\star}_k$ the resulting local plans,
which can be then sent to all agents.

\subsubsection{Static and Unknown Tasks}
For the second case, the number and location of the subtasks are unknown or uncertain,
but the subtasks remain static and immobile during execution.
For instance, for the task of ``search and rescue'', the exact number of victims
within the region is unknown and can only be determined during online execution.
Thus, a simultaneous exploration and coordination (SEC) method is proposed.
To begin with, a collaborative exploration strategy is adopted
for the subteam~$\mathcal{N}^\ell_k$ to explore the region~$S^\ell_k$
for potential subtasks, e.g., frontiers-based~\cite{holz2010evaluating}
and sampling-based~\cite{duberg2022ufoexplorer}.
Without loss of generality, the set of exploration subtasks
at time~$t>0$ is associated with the points to visit in the region,
denoted by~$\mathcal{J}^{\texttt{e}}_{t}$.
Moreover, numerous collaborative subtasks in~$\mathcal{J}^\ell_k$
are detected at time~$t$ along with its location and the required number of agents,
denoted by~$\mathcal{J}^{\texttt{c}}_{t}$.
Consequently, the set of known subtasks
that has not been fulfilled is denoted by~$\mathcal{J}_{t}\triangleq
\mathcal{J}^{\texttt{e}}_{t}\cup \mathcal{J}^{\texttt{c}}_{t}$.
Since the subtasks in~$\mathcal{J}_{t}$ are constantly changing,
a rolling assignment algorithm similar to
Alg.~\ref{alg:task} is adopted, i.e.,
to assign subtasks within~$\mathcal{J}_{t}$ in small batches
via the optimal algorithm described in the first case.
This procedure continues until the region is fully explored and
all detected subtasks are completed.

\subsubsection{Dynamic and Known Tasks}
For the third case, the total number of the subtasks and their locations are known,
but the subtasks are dynamic and mobile during execution.
For instance, the task of ``collaborative capture'' often requires the agents to
form subteams in order to surround and capture numerous moving targets.
In this case, the previous two strategies are not suitable as the motion of subtasks
would quickly render the current plans highly suboptimal or even infeasible.
Thus, a dynamic coalition formation (DCF) method is adopted for this case,
as proposed in our earlier work~\cite{chen2024accelerated}.
Particularly, each agent~$i\in \mathcal{N}^\ell_k$ only decides the next subtask
to perform, along with other agents as a coalition, i.e.,
$i\in \mathcal{N}_{j}\in \widehat{\mathcal{N}}_t$,
where~$\widehat{\mathcal{N}}_t\triangleq \{\mathcal{N}_j,\, j\in \mathcal{J}^\ell_k\}$
is the coalition scheme with all coalitions at time~$t>0$;
it holds that~$\mathcal{N}_{j_1}\cap \mathcal{N}_{j_2}=\emptyset$
and $\cup_{j\in \mathcal{J}^\ell_k}\mathcal{N}_j\subseteq \mathcal{N}^\ell_k$.
It has been proven in~\cite{chen2024accelerated} that the DCF method converges to a K-serial
stable (KSS) coalition scheme after a finite number of distributed coordination.
The detailed algorithm is omitted here due to limited space.
Afterwards, the agents would complete the assigned subtask as coalitions
and the coalition scheme is updated each time a subtask is completed.

\begin{remark}\label{rm:case}
  The case of dynamic and unknown tasks is not considered since:
  (I) Without knowing the total number and locations
of subtasks, it is difficult to determine whether the current task is completed;
(II) A combination of the strategy for the second and third cases above
would suffice.
    \hfill $\blacksquare$
\end{remark}

%% file: contents/overall.tex
\subsection{Overall Framework}\label{subsec:overall}

\subsubsection{Online Execution and Adaptation}\label{subsubsec:execute}
Initially at~$t=0$,
given the initially-known workspace and mission descriptions,
the missions are decomposed into tasks,
based on which the set of local teams are formed
as~$\overline{\mathcal{N}}$.
These tasks are assigned to the teams with a given horizon~$H$
and a redundancy~$\rho$ via Alg.~\ref{alg:task},
yielding the local plan~$\xi_k$
for each team~$\mathcal{N}_k \in \overline{\mathcal{N}}$.
Afterwards, the teams start executing
the task~$(S^\ell_k,\omega^\ell_k)\in \xi_k$
by navigating to the desired region~$S^\ell_k$
and performing the task~$\omega^\ell_k$.
Depending on the exact type of task~$\omega^\ell_k$,
the set of subtasks~$\mathcal{J}^\ell_k$ contained within~$\omega^\ell_k$
is executed by team~$\mathcal{N}_k$
following one of the three local coordination strategies
to derive local action plans~$\{\tau_i\}$.
Note that all teams are executed concurrently,
and all agents within the same team are also acting in parallel.

The conditions for replanning are designed as follows:
(I) If more than half of the assigned~$H$ tasks are accomplished;
(II) If new missions are specified;
or (III) If the local coordination of certain subteams returns infeasible.
During replanning, the task assignment and subteam formation are updated
by calling Alg.~\ref{alg:task} given the current system state.
However, the tasks that are currently being executed
can not be preempted,
which is essential when there are significantly more tasks
than the number of subteams.

\subsubsection{Complexity Analysis}\label{subsubsec:complexity}
In each iteration of Alg.~\ref{alg:task}, since $H$ tasks are assigned,
the complexity reaches~$\mathcal{O}(H2^H)$.
During subteam formation, $H\cdot N$ Boolean variables are introduced
to indicate the membership of agents in subteams.
Regarding different types of tasks,
$|\mathcal{N}_k|(J^\ell_k)^2$ integer variables are introduced to solve the MILP
for the static and known tasks;
the planning complexity for the static and unknown tasks
is~$\mathcal{O}((J^\ell_k)^3)$, similar to Alg.~\ref{alg:task};
and the complexity for the dynamic and unknown tasks
is $\mathcal{O}((N_{\omega_k} J^\ell_k)(2J^\ell_k + N_{\omega_k}) |\mathcal{N}_k|)$,
where~$N_{\omega_k}$ is the upper bound of agent number
for each coalition~\cite{chen2024accelerated}.

\begin{figure}[t]
  \centering
  \includegraphics[width=0.96\hsize]{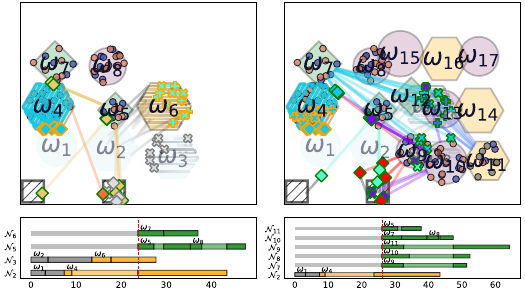}
    \vspace{-0.15in}
    \caption{Snapshots of simulation at~$t=23s$ (\textbf{Left})
      and $t=62s$ (\textbf{Right}) when new missions are released
    and replanning occurs.}\label{fig:sim-1}
  \vspace{-0.2in}
\end{figure}

%% file: contents/experiment.tex
\section{Numerical Experiments} \label{sec:experiments}
To numerical validations, the proposed method is implemented in \texttt{Python3}
and tested on a laptop with an Intel Core i5-12500H CPU.
The solver~\texttt{GLOP}~\cite{glop} is adopted for integer optimization.
Simulation videos can be found in the supplementary files.

\subsection{System Description}\label{subsec:description}
As shown in Fig.~\ref{fig:sim-1}, the simulated fleet consists
of~$N = 80$ heterogeneous agents in an open environment with map
size~$30 \text{m} \times 25 \text{m}$.
The agents fall into~$3$ kinds with varying capabilities:
20 Type-A agents capable of perception and delivery,
30 Type-B agents capable of perception and grasping,
and 30 Type-C agents capable of delivery and grasping.
Initially, the agents are distributed evenly at two bases.
All agents adhere to first-order dynamics and have a maximum speed
of~$1.5 \text{m/s}$ in simulation.

Moreover, there are~${|{\boldsymbol{\varphi}}_t|} = 4$ missions released
at random time instants,
with an average interval~$\mu = 30s$ with a standard deviation of~$\sigma = 10s$.
The sc-LTL missions follow a template
of~$\varphi_i=\Diamond (\varphi_{\texttt{del}}\wedge \Diamond \varphi_{\texttt{surv}})
\wedge (\neg \varphi_{\texttt{cap}}\mathcal{U} \varphi_{\texttt{surv}})$,
where the~$3$ types of tasks are:
``delivery'' task, requiring delivery or grasping for~$2$ different subtasks;
``surveillance'' task, requiring perception;
and ``dynamic capture'' task, requiring delivery or grasping for~$2$ different subtasks.
Delivery tasks have in average~$13$ subtasks,
and~$15$ subtasks for surveillance tasks with a probability of~$0.5$ to be unknown.
The capture tasks have around~$17$ dynamic targets with speed~$0.5 \text{m/s}$ inside the region.
The planning horizon is set to~$H = 6$ and replanning conditions follow the \ref{subsec:overall}.

\begin{figure}[t]
    \centering
    \includegraphics[width=0.97\hsize]{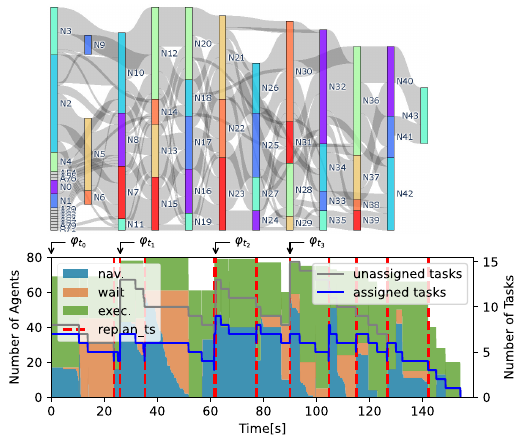}
    \vspace{-0.1in}
    \caption{\textbf{Top}: the number of subteams and their
      composition;
      \textbf{Bottom}: the status of agents in different modes
      and the number of tasks.
    }\label{fig:subteam}
    \vspace{-0.2in}
  \end{figure}

\subsection{Results}\label{subsec:results}
As shown in Fig.~\ref{fig:overall} and~\ref{fig:sim-1},
the first mission is known initially,
it takes~$0.2s$ for the method from~\cite{liu2024time}
to compute its posets, yielding 8 tasks in total.
Among these tasks, $8$ pairs follow the ``$\preceq$'' relation
and~$1$ for the ``$\simeq$'' relation.
Moreover, it takes~$1.1s$ for Alg.~\ref{alg:task} to determine that subteams are required
with a predicted makespan of~$41s$.
The subteams consist of maximum~$25$ agents and minimum~$4$ agents.
For the first task of each subteam,
the completion time is in average $15.9s$ and in total $51$ subtasks are detected.
The average planning time for~$3$ types of tasks is $0.3s$, $0.2s$ and $0.5s$, respectively.
After~$3$ tasks are completed at~$t=23.6s$, replanning is triggered,
yielding~$3$ new and~$1$ old subteams for the remaining tasks.
At~$t=26s$, another mission is released, which contains~$9$ tasks with~$10$ relations.
Thus, replanning is triggered, yielding $3$ subteams and a predicted makespan of~$48.1s$,
with a planning time of~$0.13s$.
The procedure continues with new missions released at~$62s$ and~$90s$, which leads to
a \textbf{total number of~$30$ tasks and~$492$ subtasks}.
The complete mission is accomplished at~$154.3s$, during
which~$12$ replannings are triggered.
As shown in Fig.~\ref{fig:sim-1} and~\ref{fig:subteam},
the agents switch among navigation, waiting for collaboration,
and task execution,
of which the trajectories depend heavily on the type of tasks.
Fig.~\ref{fig:subteam} shows that the number of subteams and their composition
change \emph{dynamically} online, along with the number of tasks.

\subsection{Comparisons}\label{subsec:comparisons}
The proposed method is compared against \textbf{{six}} baselines:
(I) \textbf{C-MILP-1}, where a complete MILP is formulated for
all agents~$\mathcal{N}$ and tasks~$\overline{\Omega}_t$
similar to~\cite{torreno2017cooperative,luo2021temporal},
i.e., without the subteam formation in Alg.~\ref{alg:task};
(II) \textbf{C-MILP-2}, which formulates a complete MILP directly
for all agents~$\mathcal{N}$ and subtasks~$\{\mathcal{J}^\ell_k\}$,
i.e., without the hierarchical scheme;
(III) \textbf{SAMP-1}, where a sampling-based planner from~\cite{kantaros2020stylus}
is adopted for all agents and tasks;
(IV) \textbf{SAMP-2}, which applies the sampling-based planner directly to subtasks;
(V) \textbf{Inf-H}, which is the same as our method but with a infinite
horizon~$H$, i.e., all known tasks are assigned in Alg.~\ref{alg:task};
(VI) \textbf{Greedy}, which assigns maximum one task to each subteam,
i.e., without the horizon~$H$.
As summarized in Table~\ref{table:comparasion}.
the proposed method excels at almost all metrics including response time,
planning time and success rate,
compared with \textbf{C-MILP-1,2} and \textbf{SAMP-1,2}.
Particularly, via the proposed hierarchical solution,
the planning time is~$50$ times lower than~\textbf{C-MILP-1} and \textbf{SAMP-1}
that directly assign agents to tasks.
Moreover, the methods~\textbf{C-MILP-2} and \textbf{SAMP-2} often leads to
unsuccessful executions without considering uncertainties in subtasks.
Lastly, the maximum planning time for~\textbf{Inf-H} can be prohibitively
long~($\geq 15$\text{min}),
while~\textbf{Greedy} deploys $65\%$ more agents in travelling than our methods.

\begin{table}[t]
  \begin{center}
  \begin{threeparttable}
    \caption{Comparison with Baselines}\label{table:comparasion}
    \vspace{-0.05in}
    \setlength{\tabcolsep}{0.60\tabcolsep}
    \centering
    \renewcommand{\arraystretch}{1.1}
      \begin{tabular}{c c c |c| c c c c}
      \toprule
      \multicolumn{3}{c|}{Env.}& \multirow{2}{*}{\makecell{Methods}} & \multirow{2}{*}{\makecell{Resp. \\ Time[s]}} & \multirow{2}{*}{\makecell{Ave/Max \\ Plan Time[s]}} & \multirow{2}{*}{\makecell{T/W/X \\ Agents}} & \multirow{2}{*}{\makecell{Succ. \\ Rate[\%]}} \\
      \cline{1-3}
      N & M & J & & & & & \\
      \midrule
      \multirow{7}{*}{80} & \multirow{7}{*}{30} & \multirow{7}{*}{492} & \textbf{Ours} & {62.7} & {1.4/4.9} & {20/11/37} & {100} \\
      \multirow{7}{*}{} & \multirow{7}{*}{} & \multirow{7}{*}{} & \text{C-MILP-1} & {88.5} & {100/192} & {16/6/26} & {100} \\
      \multirow{7}{*}{} & \multirow{7}{*}{} & \multirow{7}{*}{} & \text{C-MILP-2} & {28.3} & {19/64} & {12/10/25} & {86} \\
      \multirow{7}{*}{} & \multirow{7}{*}{} & \multirow{7}{*}{} & \text{SAMP-1} & {108.3} & {54/92} & {19/8/24} & {100} \\
      \multirow{7}{*}{} & \multirow{7}{*}{} & \multirow{7}{*}{} & \text{SAMP-2} & {40.7} & {5.1/16} & {23/5/20} & {85} \\
      \multirow{7}{*}{} & \multirow{7}{*}{} & \multirow{7}{*}{} & \text{Inf-H} & {77.5} & {$> 15 \text{min}$} & {24/17/31} & {100} \\
      \multirow{7}{*}{} & \multirow{7}{*}{} & \multirow{7}{*}{} & \text{Greedy} & {112.1} & {0.3/0.6} & {33/9/30} & {100} \\
      \bottomrule
    \end{tabular}
  \end{threeparttable}
 \end{center}
  \vspace{-3mm}
  \end{table}

 \begin{table}[t]
  \begin{center}
  \begin{threeparttable}
    \caption{SCALABILITY ANALYSIS}\label{table:scalability}
    \vspace{-0.05in}
    \setlength{\tabcolsep}{0.7\tabcolsep}
    \centering
    \renewcommand{\arraystretch}{1.1}
    \begin{tabular}{c c c c | c c c c}
      \toprule
      \multicolumn{4}{c|}{Env.} & \multirow{2}{*}{\makecell{Resp. \\ Time[s]}} & \multirow{2}{*}{\makecell{Ave/Max \\ Plan Time [s]}} & \multirow{2}{*}{\makecell{T/W/X \\ Agents}} & \multirow{2}{*}{\makecell{Succ. \\ Rate[\%]}}\\
      \cline{1-4}
      N & M & J & $\alpha$ & & & \\
      \midrule
      \multirow{2}{*}{120} & \multirow{2}{*}{50} & \multirow{2}{*}{824} & 0.05 & {95} & {1.6/4.6} & {33/10/39} & {100} \\
      \multirow{2}{*}{} & \multirow{2}{*}{} & \multirow{2}{*}{} & 0.1 & {102} & {1.7/4.9} & {26/9/35} & {100}\\
      \midrule
      \multirow{2}{*}{150} & \multirow{2}{*}{80} & \multirow{2}{*}{1319} & 0.05 & {153} & {1.8/5.1} & {37/12/37} & {100}\\
      \multirow{2}{*}{} & \multirow{2}{*}{} & \multirow{2}{*}{} & 0.1 & {165} & {2.1/6.2} & {12/11/34} & {97}\\
      \bottomrule
    \end{tabular}
  \end{threeparttable}
 \end{center}
  \vspace{-0.3in}
  \end{table}

  \subsection{Generalization}\label{subsec:scalability}
  For further validation, the fleet size is further increased and the agents can fail
  with a probability of~$\alpha$.
(I) \emph{\textbf{Scalability}}:
As summarized in Table~\ref{table:scalability},
when the fleet size is increased to~$120$ and $150$, while the number of tasks
to~$50$ and $80$,
at $\alpha=0.05$,
the average planning time is increased by $12.5\%$ from $1.6s$ to $1.8s$,
while the maximum planning time increases from $4.6s$ to $5.1$.
Moreover, the average response time decreases from $95$ to $153$,
as the average number of deployed agents increases from $82$ to $86$.
(II) \emph{\textbf{Failure Recovery}}:
When $\alpha=0.05$, the success rate remains~$100\%$ for~$150$ agents even
when the task number reaches~$80$.
However, the success rate drops to~$97\%$ when the failure rate reaches~$0.1$,
due the limits of remaining agent capacities.

%% file: contents/conclusion.tex
\section{Conclusion} \label{sec:conclusion}
This work proposes a hierarchical coordination framework (HULK)
that combines the global task assignment and the local subtask coordination,
under continual and uncertain collaborative tasks.
Future work includes human interaction and motion constraints.

%% file: contents/references.bib
@article{guo2016task,
  title={Task and motion coordination for heterogeneous multiagent systems with loosely coupled local tasks},
  author={Guo, Meng and Dimarogonas, Dimos V},
  journal={IEEE Transactions on Automation Science and Engineering},
  volume={14},
  number={2},
  pages={797--808},
  year={2016},
  publisher={IEEE}
}

@article{schillinger2018simultaneous,
  title={Simultaneous task allocation and planning for temporal logic goals in heterogeneous multi-robot systems},
  author={Schillinger, Philipp and B{\"u}rger, Mathias and Dimarogonas, Dimos V},
  journal={The international journal of robotics research},
  volume={37},
  number={7},
  pages={818--838},
  year={2018},
  publisher={Sage Publications Sage UK: London, England}
}

@article{kantaros2020stylus,
  title={Stylus*: A temporal logic optimal control synthesis algorithm for large-scale multi-robot systems},
  author={Kantaros, Yiannis and Zavlanos, Michael M},
  journal={The International Journal of Robotics Research},
  volume={39},
  number={7},
  pages={812--836},
  year={2020},
  publisher={SAGE Publications Sage UK: London, England}
}

@article{luo2021temporal,
  title={Temporal Logic Task Allocation in Heterogeneous Multi-Robot Systems},
  author={Luo, Xusheng and Zavlanos, Michael M},
  journal={arXiv preprint arXiv:2101.05694},
  year={2021}
}

@book{baier2008principles,
  title={Principles of model checking},
  author={Baier, Christel and Katoen, Joost-Pieter},
  year={2008},
  publisher={MIT press}
}

@article{sahin2019multirobot,
  title={Multirobot coordination with counting temporal logics},
  author={Sahin, Yunus Emre and Nilsson, Petter and Ozay, Necmiye},
  journal={IEEE Transactions on Robotics},
  volume={36},
  number={4},
  pages={1189--1206},
  year={2019},
  publisher={IEEE}
}

@inproceedings{jones2019scratchs,
  title={ScRATCHS: Scalable and robust algorithms for task-based coordination from high-level specifications},
  author={Jones, Austin M and Leahy, Kevin and Vasile, Cristian and Sadraddini, Sadra and Serlin, Zachary and Tron, Roberto and Belta, Calin},
  booktitle={Proc. Int. Symp. Robot. Res.},
  pages={1--16},
  year={2019}
}

@book{hoos2004stochastic,
  title={Stochastic local search: Foundations and applications},
  author={Hoos, Holger H and St{\"u}tzle, Thomas},
  year={2004},
  publisher={Elsevier}
}

@article{toth2002overview,
  title={An overview of vehicle routing problems},
  author={Toth, Paolo and Vigo, Daniele},
  journal={The vehicle routing problem},
  pages={1--26},
  year={2002},
  publisher={SIAM}
}

@inproceedings{varava2017herding,
  title={Herding by Caging: a Topological Approach towards Guiding Moving Agents via Mobile Robots.},
  author={Varava, Anastasiia and Hang, Kaiyu and Kragic, Danica and Pokorny, Florian T},
  booktitle={Robotics: Science and Systems},
  pages={696--700},
  year={2017}
}

@inproceedings{cliff2015online,
  title={Online localization of radio-tagged wildlife with an autonomous aerial robot system},
  author={Cliff, Oliver M and Fitch, Robert and Sukkarieh, Salah and Saunders, Debra L and Heinsohn, Robert},
  booktitle={Robotics: Science and Systems},
  year={2015}
}

@article{arai2002advances,
  title={Advances in multi-robot systems},
  author={Arai, Tamio and Pagello, Enrico and Parker, Lynne E and others},
  journal={IEEE Transactions on robotics and automation},
  volume={18},
  number={5},
  pages={655--661},
  year={2002},
  publisher={Citeseer}
}

@inproceedings{fink2008multi,
  title={Multi-robot manipulation via caging in environments with obstacles},
  author={Fink, Jonathan and Hsieh, M Ani and Kumar, Vijay},
  booktitle={2008 IEEE International Conference on Robotics and Automation},
  pages={1471--1476},
  year={2008},
  organization={IEEE}
}

@article{guo2015multi,
  title={Multi-agent plan reconfiguration under local LTL specifications},
  author={Guo, Meng and Dimarogonas, Dimos V},
  journal={The International Journal of Robotics Research},
  volume={34},
  number={2},
  pages={218--235},
  year={2015},
  publisher={SAGE Publications Sage UK: London, England}
}

@article{tumova2016multi,
  title={Multi-agent planning under local LTL specifications and event-based synchronization},
  author={Tumova, Jana and Dimarogonas, Dimos V},
  journal={Automatica},
  volume={70},
  pages={239--248},
  year={2016},
  publisher={Elsevier}
}

@article{torreno2017cooperative,
  title={Cooperative multi-agent planning: A survey},
  author={Torre{\~n}o, Alejandro and Onaindia, Eva and Komenda, Anton{\'\i}n and {\v{S}}tolba, Michal},
  journal={ACM Computing Surveys (CSUR)},
  volume={50},
  number={6},
  pages={1--32},
  year={2017},
  publisher={ACM New York, NY, USA}
}

@inproceedings{gini2017multi,
  title={Multi-robot allocation of tasks with temporal and ordering constraints},
  author={Gini, Maria},
  booktitle={AAAI Conference on Artificial Intelligence},
  year={2017}
}

@inproceedings{boerkoel2013distributed,
  title={Distributed algorithms for incrementally maintaining multiagent simple temporal networks},
  author={Boerkoel Jr, James C and Planken, L{\'e}on R and Wilcox, Ronald J and Shah, Julie A},
  booktitle={International Conference on Automated Planning and Scheduling},
  year={2013}
}

@article{luo2015distributed,
  title={Distributed algorithms for multirobot task assignment with task deadline constraints},
  author={Luo, Lingzhi and Chakraborty, Nilanjan and Sycara, Katia},
  journal={IEEE Transactions on Automation Science and Engineering},
  volume={12},
  number={3},
  pages={876--888},
  year={2015},
  publisher={IEEE}
}

@article{fukasawa2006robust,
  title={Robust branch-and-cut-and-price for the capacitated vehicle routing problem},
  author={Fukasawa, Ricardo and Longo, Humberto and Lysgaard, Jens and De Arag{\~a}o, Marcus Poggi and Reis, Marcelo and Uchoa, Eduardo and Werneck, Renato F},
  journal={Mathematical programming},
  volume={106},
  number={3},
  pages={491--511},
  year={2006},
  publisher={Springer}
}

@article{khamis2015multi,
  title={Multi-robot task allocation: A review of the state-of-the-art},
  author={Khamis, Alaa and Hussein, Ahmed and Elmogy, Ahmed},
  journal={Cooperative Robots and Sensor Networks 2015},
  pages={31--51},
  year={2015},
  publisher={Springer}
}

@inproceedings{nunes2015multi,
  title={Multi-robot auctions for allocation of tasks with temporal constraints},
  author={Nunes, Ernesto and Gini, Maria},
  booktitle={Proceedings of the AAAI Conference on Artificial Intelligence},
  volume={29},
  number={1},
  year={2015}
}

@article{lahijanian2011temporal,
  title={Temporal logic motion planning and control with probabilistic satisfaction guarantees},
  author={Lahijanian, Morteza and Andersson, Sean B and Belta, Calin},
  journal={IEEE Transactions on Robotics},
  volume={28},
  number={2},
  pages={396--409},
  year={2011},
  publisher={IEEE}
}

@article{brucker1994branch,
  title={A branch and bound algorithm for the job-shop scheduling problem},
  author={Brucker, Peter and Jurisch, Bernd and Sievers, Bernd},
  journal={Discrete applied mathematics},
  volume={49},
  number={1-3},
  pages={107--127},
  year={1994},
  publisher={Elsevier}
}

@article{massin2017coalition,
  title={A coalition formation game for distributed node clustering in mobile ad hoc networks},
  author={Massin, Rapha{\"e}l and Le Martret, Christophe J and Ciblat, Philippe},
  journal={IEEE Transactions on Wireless Communications},
  volume={16},
  number={6},
  pages={3940--3952},
  year={2017},
  publisher={IEEE}
}

@article{apt2009generic,
  title={A generic approach to coalition formation},
  author={Apt, Krzysztof R and Witzel, Andreas},
  journal={International game theory review},
  volume={11},
  number={03},
  pages={347--367},
  year={2009},
  publisher={World Scientific}
}

@inproceedings{menghi2018multi,
  title={Multi-robot LTL planning under uncertainty},
  author={Menghi, Claudio and Garcia, Sergio and Pelliccione, Patrizio and Tumova, Jana},
  booktitle={International Symposium on Formal Methods},
  pages={399--417},
  year={2018},
  organization={Springer}
}

@inproceedings{kantaros2018distributed,
  title={Distributed optimal control synthesis for multi-robot systems under global temporal tasks},
  author={Kantaros, Yiannis and Zavlanos, Michael M},
  booktitle={2018 ACM/IEEE 9th International Conference on Cyber-Physical Systems (ICCPS)},
  pages={162--173},
  year={2018},
  organization={IEEE}
}

@phdthesis{schillinger2019specification,
  title={Specification Decomposition and Formal Behavior Generation in Multi-Robot Systems},
  author={Schillinger, Philipp},
  year={2019},
  school={KTH Royal Institute of Technology}
}

@misc{glop,
author={Google Linear Optimization Solver},
howpublished={\texttt{\url{https://developers.google.com/optimization/lp}}}
}

@ARTICLE{chen2024accelerated,
  author={Chen, Junfeng and Tang, Zili and Guo, Meng},
  journal={IEEE Robotics and Automation Letters}, 
  title={Accelerated K-Serial Stable Coalition for Dynamic Capture and Resource Defense}, 
  year={2024},
  volume={9},
  number={1},
  pages={443-450}
  }

@article{liu2024fast,
  title={Fast and Adaptive Multi-Agent Planning under Collaborative Temporal Logic Tasks via Poset Products},
  author={Liu, Zesen and Guo, Meng and Bao, Weimin and Li, Zhongkui},
  journal={Research},
  volume={7},
  pages={0337},
  year={2024},
  publisher={AAAS}
}

@article{liu2024time,
  title={Time minimization and online synchronization for multi-agent systems under collaborative temporal logic tasks},
  author={Liu, Zesen and Guo, Meng and Li, Zhongkui},
  journal={Automatica},
  volume={159},
  pages={111377},
  year={2024},
  publisher={Elsevier}
}

@inproceedings{holz2010evaluating,
  title={Evaluating the efficiency of frontier-based exploration strategies},
  author={Holz, Dirk and Basilico, Nicola and Amigoni, Francesco and Behnke, Sven},
  booktitle={International Symposium on Robotics},
  pages={1--8},
  year={2010},
  organization={VDE}
}

@article{duberg2022ufoexplorer,
  title={Ufoexplorer: Fast and scalable sampling-based exploration with a graph-based planning structure},
  author={Duberg, Daniel and Jensfelt, Patric},
  journal={IEEE Robotics and Automation Letters},
  volume={7},
  number={2},
  pages={2487--2494},
  year={2022},
  publisher={IEEE}
}
